# Characterising and Analysing Retrieval-Augmented Generation Techniques

Emmanuel Igbekele[1, 2] and Olawande Daramola [1]

[1]Department of Informatics, University of Pretoria, Pretoria 0083, South Africa
emmanuel.igbekele@up.ac.za, wande.daramola@up.ac.za
[2]Department of Computer Science, Landmark University, Kwara State, Nigeria
igbekele.emmanuel@lmu.edu.ng

**Abstract.** The advent of Large Language Models (LLMs) across many application domains has increased the need for Retrieval-Augmented Generation (RAG) to reduce hallucinations and improve LLM performance. Thus far, several RAG techniques with impressive features and capabilities have been proposed in the literature, making it a herculean task to select the most appropriate RAG techniques for specific problem scenarios. Thus, we propose a conceptual framework comprising main dimensions and sub-dimensions for characterising (defining "what it is") and analysing (defining "how well it does") existing RAG techniques. The conceptual framework was developed through a qualitative analysis of the attributes of RAG techniques reported in the literature. It can serve as a pre-empirical filtering mechanism for researchers and practitioners in selecting the highest-fit RAG candidates that meet specific user and task requirements before empirical benchmarking of RAG techniques, which can be computationally and financially expensive.

## 1 Introduction

Recently, new AI models, such as Generative Artificial Intelligence (GenAI), have been increasingly deployed to assist with data-driven learning and probabilistic modelling, produce new explainable content such as text, images, or audio, and make decisions or solve recurring problems.

GenAI architectures are usually based on large language models (LLMs). A large language model (LLM) possesses natural language processing (NLP) capabilities for text generation and language comprehension. However, problems with factual accuracy, currentness of information, and hallucinations still exist [1]. So far, Retrieval-Augmented Generation (RAG) has proven to be one of the most effective mechanisms for addressing the limitations of traditional LLMs. RAG enables the real-time integration of knowledge retrieved from external sources to aid LLMs' language generation capabilities [2].

Knowledge presentation for RAG takes various forms, each with its own benefits. Organisations increasingly adopt RAG to build AI Software-as-a-Service (SaaS) products and features. Different classes and types of RAG techniques exist, with various proposals highlighting their strength and weaknesses. Existing comparative studies of RAG techniques tend to focus on limited aspects/parameters without a holistic appraisal of essential aspects [3,4,5]. Also, exhaustive empirical comparative analysis of RAG techniques using standard RAG benchmarks with different metrics is computationally and financially expensive [6]. Thus, selecting the most appropriate RAG technique that fits a specific task can be challenging for researchers and practitioners [7].

In this paper, we propose a conceptual framework that enables the characterisation (defining what it is) and analysis (assessing how well it performs) of RAG techniques to aid decision-making in selecting the most appropriate RAG technique for a specific task. The conceptual framework was derived from a qualitative review of the literature, with a particular focus on previously published review, survey, and comparative review papers. We analysed the features of RAG techniques as profiled in the literature to identify the main dimensions and sub-dimensions of the proposed framework. As a contribution, this study presents a conceptual framework for characterising RAG techniques in terms of their design attributes and for analysing their functional and operational capabilities, providing a robust basis for decision-making when selecting RAG techniques that fit specific task/user requirements. It offers a more comprehensive framework consisting of 11 dimensions and subdimensions that surpasses the scope covered by previous efforts, and provides a pre-empirical filtering mechanism for selecting the most high-fit RAG techniques that meet specific user and task requirements before empirical validation.

The rest of this paper is organised as follows. Section 2 presents the theoretical background and a review of related work. In Section 3, we describe the methodology adopted to develop the framework. The proposed framework is described in Section 4, and the paper concludes in Section 5 with a brief note and an overview of future research prospects.

## 2 Background and Related Work

This section presents an overview of RAG and a review of related work.

### 2.1 Retrieval Augmented Generation (RAG)

RAG is an advanced AI architecture designed to enhance LLM performance [8]. RAG is made possible by establishing a dynamic connection between the chosen model(s) and external, authoritative knowledge bases during text generation [1]. With RAG, users can directly interact with data repositories, thereby opening up new kinds of experiences, implying that the applications of RAG are weighted much more heavily than the number of available datasets [2, 9, 10]. A major motivation for RAG has been a rise in demand for context-relevant and accurately updated generated content [11]. By integrating real-time external data sources into an AI response-generation process, RAG addresses the limitations of traditional LLMs, such as outdated information and plausible yet incorrect responses, commonly referred to as "hallucinations" [12]. RAG functions essentially by combining two major components – "Retriever" and "Generator" in producing accurate and contextually appropriate responses. This "Augmentation" process allows LLMs to access up-to-date knowledge without necessarily retraining the models, thereby improving efficiency and reliability.

### 2.2 Related Work

Several authors have compared RAG techniques using different assessment parameters.

A comprehensive study of RAG was presented in [3,11] covering the evolution of RAG techniques, the core foundational concepts, and the current state of the art. The authors explored the basic architectures of a few distinct RAG techniques, extracted from 63 papers, with a special focus on retrieval and generation modalities and on how their incorporation is handled to ensure that knowledge-intensive tasks are addressed delicately. The review provided a taxonomy of RAG, highlighting key approaches in Retrieval Augmented Language Models (REALM), as well as other uses in question-response domains, summarisation, and knowledge-oriented tasks. The study focused on the basics of RAG and examined current challenges related to bias, scalability, ethics, and privacy concerns in deployment.

A RAG task categorisation method was proposed in [4] that takes user queries as input and classifies them into 4 levels based on the task's primary focus and the type of external data expected. The four levels include explicit fact queries, implicit fact queries, interpretable rationale queries, and hidden rationale queries. The study, which sought to enable LLMs to use external data more productively, split the queries into different levels, provided relevant datasets where necessary, and summarised the major challenges and the most effective techniques for addressing them. The study aimed to provide application development solutions and serve as a systematic guide to emerging challenges.

In [5], a survey of variants of Agentic RAG was presented, providing a comprehensive overview of the evolutionary principles underlying the development of RAG paradigms. The study identified seven approaches from about 100 research papers, providing a detailed taxonomy of Agentic RAG architectures, highlighting their applications in domains such as healthcare, finance, and education, and examining practical implementation strategies. Additionally, the work addressed the challenge of system scalability, ensured decision-making aligned with ethical principles, optimised performance for real-world applications, and provided detailed insights into frameworks and tools for implementing Agentic RAG.

A comprehensive survey on RAG evaluation methodologies and frameworks was provided in [13]. The authors focused on evaluation approaches that help enhance system performance, computational efficiency, safety, accuracy, and factuality. Furthermore, the authors compiled and categorised datasets used in RAG, as well as frameworks for evaluating RAG. The comparative review in [14] examined how combining retrieval mechanisms with advanced reasoning can empower LLMs to tackle increasingly complex problems. A comprehensive taxonomy was developed from 77 RAG implementations, including collaborative objectives, representative paradigms, and technical implementations, and analysing bidirectional synergy methods across multi-dimensional paradigms.

Our review of the literature reveals that, to date, no classification of RAG techniques by architectural design and operational attributes has been proposed, limiting understanding of their intrinsic composition, features, and capabilities. Thus, a more comprehensive framework that embraces multiple critical dimensions to enable rigorous characterisation and analysis of the features and capabilities of RAG techniques is needed to facilitate better decision-making when selecting RAG techniques for specific tasks.

# 3 Methodology

The methodology adopted to develop our conceptual framework is presented in Figure 1. We started with a systematic review protocol that involved formulating the research question: "How can a framework for characterising and analysing RAG techniques be developed?" A literature search and a paper selection process followed this. In all, we applied a seven-step process, which includes: i) literature search; ii) paper selection; iii) determining key themes; iv) identifying dimensions; v) identifying sub-dimensions; vi) identifying concepts and metrics of dimensions/sub-dimensions; and vii) quality assurance. The description of each step is presented next.

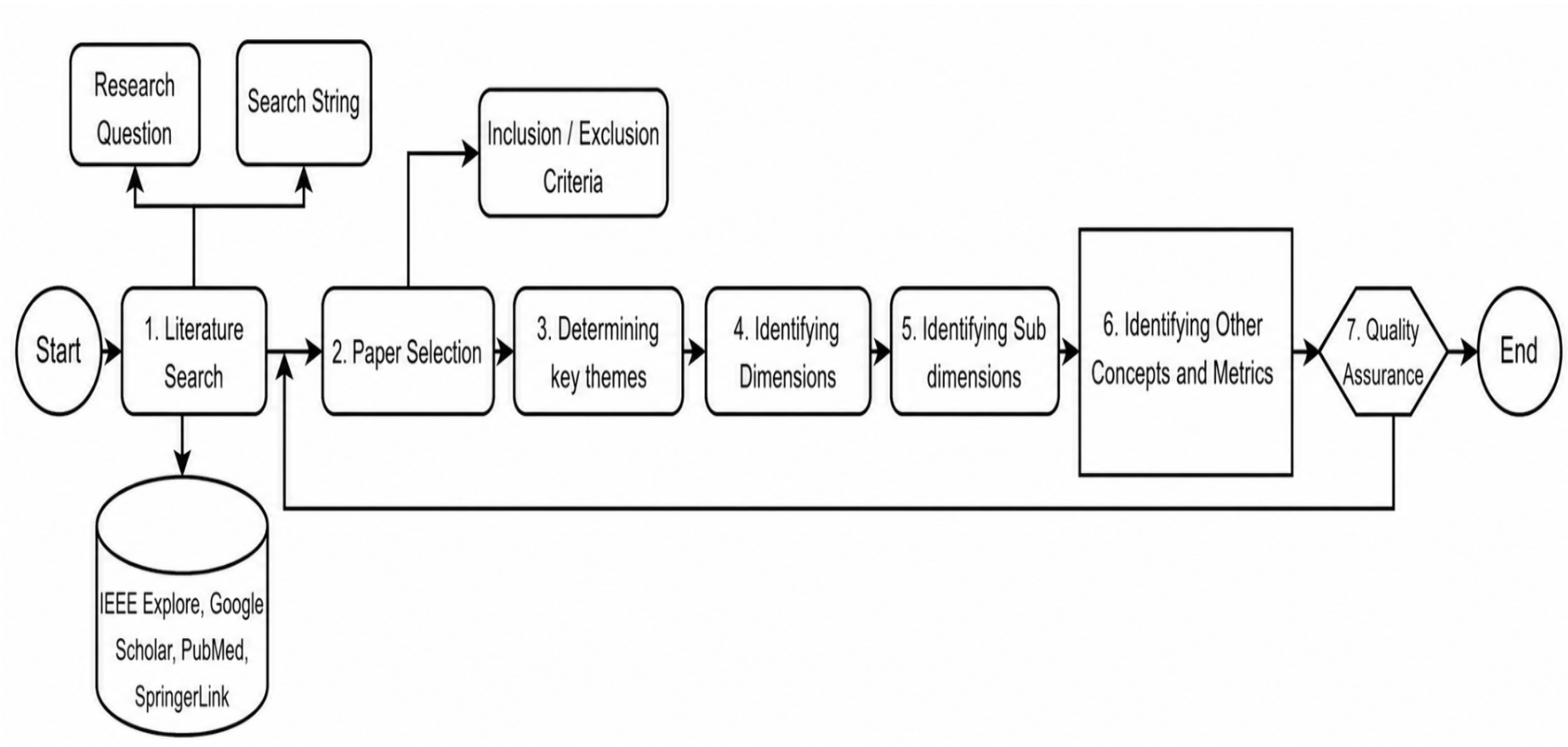

**Fig. 1.** An overview of the framework development process

i) **Literature Search**: This step involves searching for published papers in some academic databases. By using the regular Boolean operators "AND" and "OR", the following combinations of search terms were applied: "*RAG Techniques*" OR "*RAG Survey*" OR "*RAG classifications*" OR "*RAG review*" OR "*RAG models*" OR "*RAG literature review*" OR "*RAG surveys*" OR "*recent works on RAG architectures*" OR "*Application of RAG techniques in GenAI*" OR "*Retrieval Augmented Generation in GenAI*" OR "*RAG dimensions*" AND "*GenAI*" OR "*Artificial Intelligence*".

We retrieved 364 papers after applying the search string on the academic databases: Elsevier, Web of Science, Google Scholar, ScienceDirect, PubMed, IEEE Xplore, and SpringerLink (n = 364).

ii) **Paper Selection**: After initial screening, 254 papers were removed, including duplicate papers (102) and papers that are not relevant to the study (152). The remaining 110 papers were further screened to identify core RAG papers, leading to the removal of 35 papers. Thereafter, we applied the inclusion and exclusion criteria to remove 35 additional papers, resulting in the final 40 papers selected for this study (see Figure 2).

**Inclusion criteria**: Articles on RAG design/implementation, written in the English language; published between 2020 and 2025; published in journals, peer-reviewed conferences and workshops.
**Exclusion criteria**: Non-peer-reviewed publications and articles not written in the English language.

These 40 selected articles were further reviewed to identify those that provide detailed and comparative information on RAG techniques, which is pivotal to the main objective of our study. We considered surveys, systematic reviews and comparative review papers. Consequently, we identified seven (7) studies [3, 4, 5, 13, 14, 15, 16] that met this criterion. Thus, we used this as a basis to conduct a tertiary study to identify the dimensions and sub-dimensions of our conceptual framework. The other 32 papers in the pool were used as additional sources of evidence to support the identification of dimensions, sub-dimensions, and the concepts and metrics for dimensions or sub-dimensions.

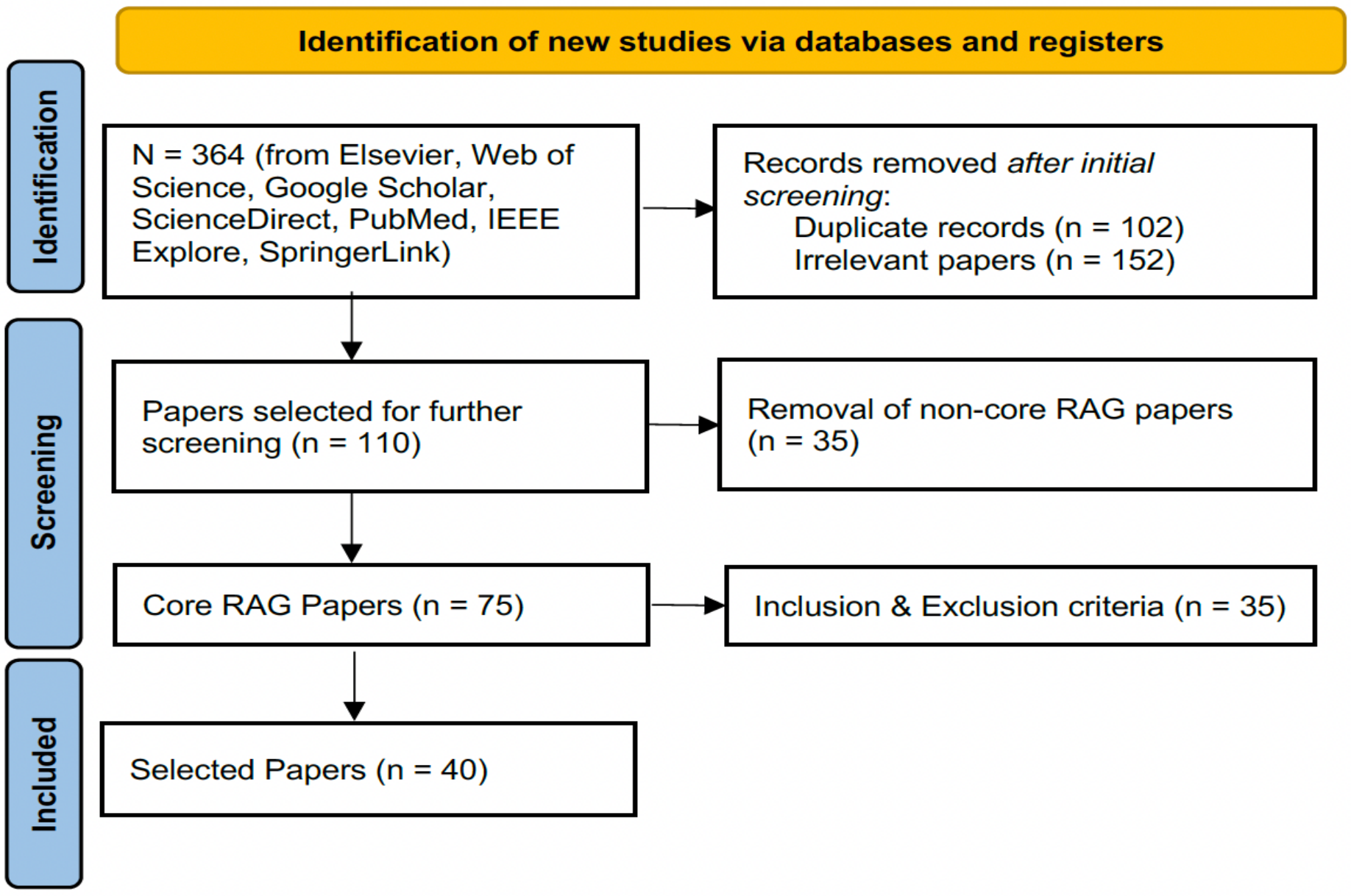

**Fig. 2.** An Overview of the Paper Selection Process

iii) **Determining Key Themes**: Using the seven selected comparative review papers for our tertiary study, we conducted a thematic analysis, which helped to identify the patterns and common themes that the authors used in conducting their comparative reviews and surveys. At the end, we identified thirty-three (33) different common themes/classes that describe the characteristics/attributes of RAG techniques (see Table 1).
iv) **Identifying Dimensions**: After reviewing the identified 33 classes, we found some classes that refer to the same concept, albeit from different perspectives. For example, the authors in [3, 4, 14] all emphasised the data retrieval mechanism but from different perspectives. Hence, the commonalities amongst the classes were streamlined into eleven (11) baseline dimensions (see Figure 3). Further appraisal of the dimensions revealed that they could be broadly categorised into characterisation and analysis dimensions. A characterisation dimension relates to the design architecture of a RAG technique, while an analysis dimension focuses on the functional and operational attributes of a RAG technique.

**Table 1.** Characterising classes of RAG techniques identified from the review

| Class | *Ref.* | Class | *Ref.* | Class | *Ref.* |
|---|---|---|---|---|---|
| Retrieval Quality | [3,17,42, 43] | Chunking Optimisation | [18,19,20,21] | Fairness | [22] |
| Trustworthiness | [13] | Interpretability | [23] | Transparenc y | [23] |
| Ethics | [3,22,23, 24,25] | Domain Application | [3,23,24] | Correction | [25,26] |
| Generative Quality | [3,26,27] | Ranking | [27] | Performance | [2,11,26,27, 28] |
| Scalability | [26,28,29 ] | Efficiency | [28,29,30] | Optimisation | [3,23,29,30] |
| Data Processing Enhancement | [29,31] | Multimodal Integration | [29,30,31] | Emerging Technologie s | [3,11,30,31, 32] |
| Operational Overhead | [13,33,34,35,36] | Developmental Cost | [33,34,35,36] | Personalisation | [3,11,37] |
| Adaptation | [3,11,38] | Robustness | [13,39,40,41] | Safety | [13,28,39,41] |
| Retrieval Relevance | [26,27,42] | Generation Enhancement | [26,44] | Task Application | [3,45] |
| Agentic Embeddings | [5,46,47,48,49,50,51,52,53] | Privacy | [3,22,54,55,56] | Coherence | [57] |
| Bias | [13,22,28,57] | Cross-lingual | [58] | Low Resource | [23,58] |

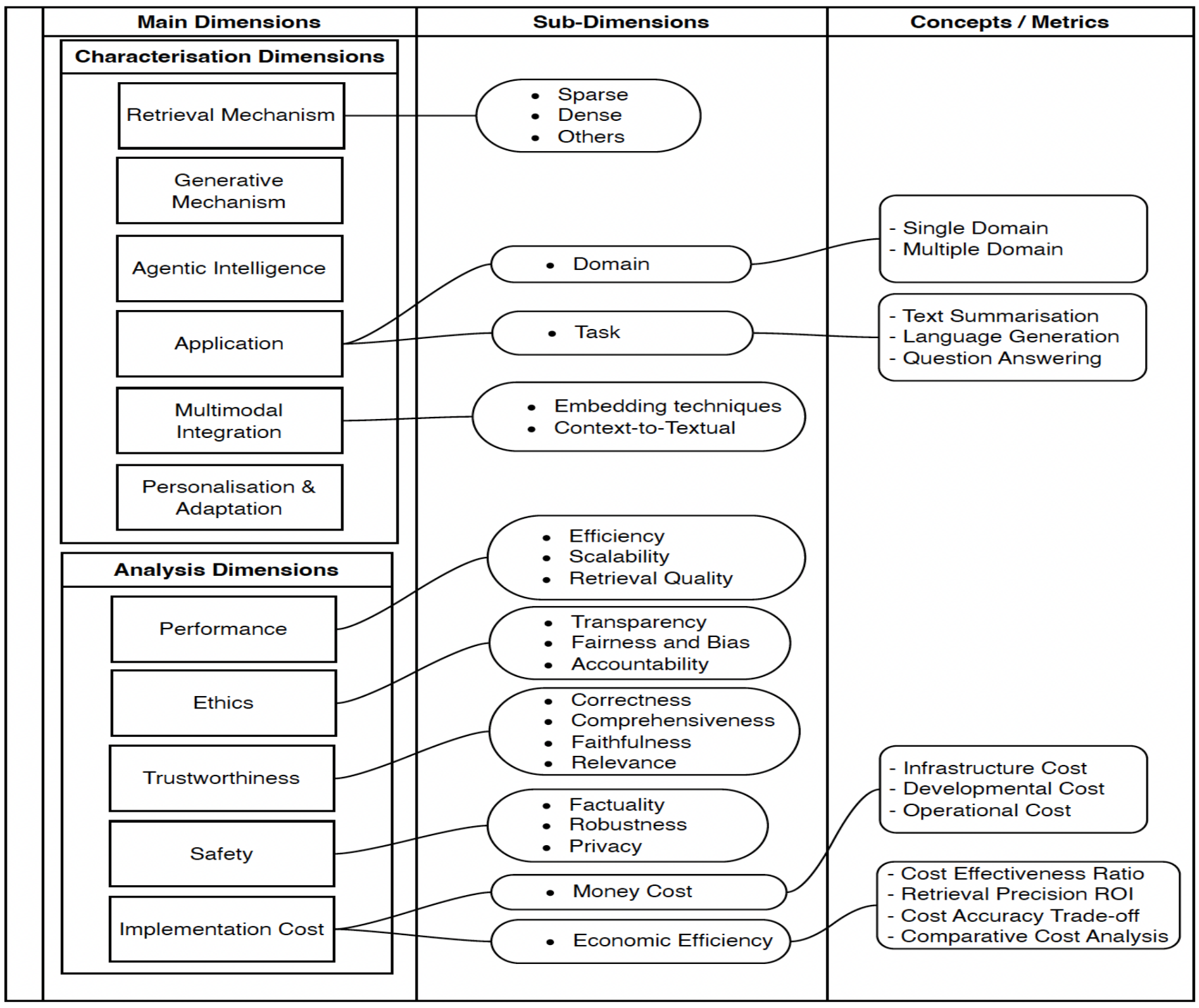

**Fig. 3.** A Conceptual framework for characterising and analysing RAG techniques.

v) **Identifying Sub-dimensions:** Furthermore, we discovered that some of the baseline dimensions have attributes that need to be explored in more detail. Hence, we created another level of abstraction called sub-dimensions.
vi) **Identifying Other Concepts and Metrics:** We observed that some of the sub-dimensions, namely, *Domain Application*, *Task Application*, *Money Cost*, and *Economic Efficiency,* have additional informational aspects which transcend the sub-dimension abstraction. Hence, we created yet a lower level of abstraction, which we called concepts/metrics, to realise a more robust and inclusive conceptual framework (see Figure 1).
vii) **Quality Assurance:** We made quality assurance an integral aspect of the framework development process through multiple rounds of rigorous review to ensure correct classification of concepts as dimensions, subdimensions and metrics of the framework. Typically, each concept proposed by one author was verified by the other, and consensus was reached before deciding whether it was classified as a dimension, subdimension, or metric.

# 4 Justification of the Composition of the Framework

This section presents the justification of the components of the framework.

## 4.1 Characterisation Dimensions

A characterisation dimension focuses on the architectural design attributes of a RAG technique. The justifications for the characterisation dimensions are presented as follows.

**i**. **Retrieval Mechanism**: Virtually all the reviewed papers presented their unique views on the basic functional components of RAG, which are *retrieval* and *generative* mechanisms. Most discussions begin with the differences between the two mechanisms and the foundational principles of LLMs. Most authors focused on *Data Retrieval Enhancement* modalities (such as indexing to establish mappings from search terms to text segments), which are classified as *Sparse*, *Dense* and *Other* kinds of retrievals [3, 4, 5, 13, 14, 15, 16].

**ii. Generative Mechanism:** The generative mechanism focuses on how a RAG technique combines retrieval with generation. The generative mechanism, as presented by [3, 4], ensures the generated text is fluent, accurate, and aligned with the original query by synthesising the retrieved information into coherent, contextually appropriate responses [3, 16, 17, 42, 43]. Common generative mechanisms include *Retrieve-then-Generate* (Retrieve docs, feed to generator), *Retrieve-and-Edit* (retrieve similar examples, edit them), *Generate-then-Retrieve* (generate query embeddings, retrieve docs), and *End-to-End* RAG [17].

**iii. Agentic Intelligence**: Agentic intelligence is an integral foundation of Agentic Retrieval-Augmented Generation (RAG) systems, enabling them to go beyond the default, static and reactive nature of traditional RAG. In [5], the five characteristics of agentic RAG systems are *Prompt Chaining*, *Routing*, *Parallelisation*, *Orchestrator-Workers* and *Evaluation-Optimiser*.

**iv. Application**: RAG systems can also be characterised by their application [3, 23]. RAG applications can be classified into domain and task categorisation. The domain subclassification can be conceptualised as a *multi-domain system* or a *single-domain system* [59]. According to [45], task classification focuses on the core functionality of the LLM within the system, including text summarisation, language generation, and context-aware question answering.

**v. Multimodal Integration**: This dimension focuses on the capacity of the RAG systems to handle different types of data. The authors in [29, 30] synthesised the integration of data in various forms, such as video, figures, tables, text, and audio, as a recurring attribute that can alter the processing of user prompts and therefore pose a challenge in output generation. The major approaches being used to handle multimodal data are *Embedding Techniques* and *Context-to-Textual* [31].

**vi. Personalisation and Adaptation**: This dimension is focused on the ability of RAG systems to enable personalised and adaptable functionalities. Currently, RAG components are usually assessed using end-to-end RAG ontology metrics, without in-depth functional decomposition, adaptation, or theoretical analysis [37]. [38] observed that LLMs can be improved through the integration of knowledge from multiple sources and also adaptive domain systems. The use of *sentiment expression* and *fine-tuning* models has been identified as a way to achieve this.

## 4.2 Analysis Dimensions

The analysis dimensions are described as follows:

**i. Performance**: According to authors in [3, 11, 23, 26, 27, 28] the performance attributes of RAG techniques are: *scalability* (the measure of capacity to handle growth in data and number of users); *efficiency* (the measure of the response time, and resource usage during retrieval); and *retrieval quality* (the measure of relevance and correctness of retrieved results). Thus, these make up the subdimensions of the performance dimension.

**ii. Ethics**: The authors in [22] identified fairness and bias as critical aspects of the ethics of RAG techniques. The need for traceability to foster understanding of the reasoning process of RAG systems was *termed "transparency"* in [23]. *Accountability* is a composite attribute encompassing traceability, error tracking, verifiability, and transparency. It provides a basis for assessing trust and ethical compliance, and for improving RAG systems [24, 25]. Thus, *Bias and Fairness*, *Transparency and Accountability* were selected as subdimensions of ethics.

**iii. Trustworthiness**: Trustworthiness defines the extent to which the RAG system can be relied on. According to [13], the critical aspects of trustworthiness are correctness (how accurate is the generated answer?), *Comprehensiveness* (how well does the answer cover the query aspects?), *Faithfulness* (how well does the answer stick to the content of retrieved sources)*, and relevance* (how well does the answer satisfy the user query?). Thus, all these aspects form the subdimensions of trustworthiness.

**iv. Safety**: Presented in [13] as external evaluation, safety is the measure of how well a RAG technique can avoid causing harm or risks. According to [32, 39, 40, 41], the core factors of safety of RAG include robustness (How well does it handle unexpected inputs, edge cases, or adversarial attacks?), factuality (how accurately does generated results reflects facts from actual sources?), privacy (how well does it protect sensitive information?) and transparency (how clear is its reasoning process, and sources of information, and limitations?). These factors make up the subdimensions of safety.

**v. Implementation Cost**: According to authors in [13, 33], irrespective of the choice that is made, cost is associated with the development of RAG techniques. According to [33], the cost perspective can be explored in terms of *monetary cost* and *economic efficiency*. The factors closely related to *monetary cost,* as highlighted by [33], are *infrastructure cost (*computing local resources for embedding generation), *developmental cost* (initial implementation, integration, and customisation expenses), and *operational cost (*human oversight, system maintenance, and up-to-date knowledge storage). According to [25, 34, 35], the factors that determine *economic efficiency* are *the cost-effectiveness ratio* and *retrieval precision ROI*, *the cost-accuracy, the cost-accuracy trade-off*, and *comparative cost analysis*. These factors make up the subdimensions and metrics of implementation cost in our conceptual framework.

## 5. Applying the Conceptual Framework

The proposed conceptual framework can be applied to comparatively assess the suitability of different RAG techniques against specific user/task requirements. Industry practitioners can use it as a filtering tool to determine the high-fit RAG candidates that should be further explored through empirical RAG benchmarking before final selection [6]. The description of how the conceptual framework can be applied is presented next.

i. **Characterisation Dimensions:** By leveraging the existing information in the RAG literature, we can rate the dimension/subdimension of each RAG technique by using a 3-category qualitative criteria checklist: √ (Yes), † (Partial), and X (No), depending on whether a dimension/subdimension is supported, partially supported or not supported by a RAG technique. If rating scores of 1 (Yes), 0.5 (Partial), and 0 (No) are applied, it will be possible to compute an overall score for different RAG techniques when measured against a specific set of tasks and user requirements (See Table 2). The qualitative ratings that were assigned to the dimensions/subdimensions were inferred from the information provided in the reviewed papers.

ii. **Analysis Dimensions:** Similarly, based on available information in the literature, we can assign a rating score to each analysis dimension/subdimension/metrics of a RAG technique based on a 5 point Likert scale: vH (Very High) = 5, H (High) = 4, M (Medium) = 3, L (Low) = 2, vL (Very Low) = 1, and Insufficient Information (-) = 0. Midpoint values were assigned to intermediate ranges, such as L-M (Low to Medium) = 2.5 and M-H (Medium to High) = 3.5, in the checklist (Table 3). Unlike the other dimensions, we did not find sufficient information from the literature to rate the subdimensions/metrics of the Ethics dimension.

### 5.1 Inter-Rater Agreement

The ratings assigned to the characterisation and analysis dimensions were a product of inter-rater agreement and consensus by the authors. Individual authors assigned ratings to the dimensions and sub-dimensions of the different RAG techniques. The agreement of ratings between the two authors was 70%. Thereafter, we discussed the differences in ratings in a meeting to arrive at the consensus ratings that we eventually assigned for each dimension and sub-dimension.

**Table 2.** Checklist of the Characterisation Dimensions for RAG techniques

"N" = Naïve RAG; "C" = Corrective RAG; "G" = Graph RAG; "A" = Agentic RAG; "M" = Multimodal RAG; "S" = Self-RAG; "V" = Vision RAG; "F" = Fusion RAG; "R" = Radio RAG; "H" = HyPA RAG.
**Rating**: √ (Yes), X (No), † (partial)

| Dimension /Subdimensions/ Metrics | | | N | C | G | A | M | S | V | F | R | H |
|---|---|---|---|---|---|---|---|---|---|---|---|---|
| *Retrieval Mechanism* | | *Sparse Retrieval* | √ | √ | √ | √ | √ | √ | √ | √ | √ | √ |
| | | *Dense Retrieval* | √ | √ | √ | √ | √ | √ | √ | √ | √ | √ |
| | | *Hybrid Retrieval* | √ | √ | √ | √ | √ | √ | √ | √ | √ | √ |
| *Generative Mechanism* | | | √ | √ | √ | √ | √ | √ | √ | √ | √ | √ |
| *Agentic Intelligence* | | | X | √ | † | √ | † | √ | † | X | X | √ |
| *Application* | *Domain* | *Single* | √ | √ | √ | √ | √ | √ | √ | √ | √ | √ |
| | | *Multiple* | X | √ | √ | √ | √ | √ | √ | √ | X | √ |
| | *Task* | *Text Summarisation* | † | √ | √ | √ | √ | √ | † | √ | √ | √ |
| | | *Language Generation* | √ | √ | √ | √ | √ | √ | √ | √ | √ | √ |
| | | *Question Answering* | √ | √ | √ | √ | √ | √ | √ | √ | √ | √ |
| *Multimodal Integration* | | *Embedding Techniques* | † | √ | √ | √ | √ | X | √ | √ | X | √ |
| | | *Context-to-Textual* | X | √ | √ | √ | √ | √ | † | √ | √ | √ |
| *Personalization & Adaptation* | | *Personalisation* | X | √ | √ | √ | † | X | † | X | X | X |
| | | *Adaptation* | X | √ | √ | √ | √ | √ | √ | √ | √ | √ |

Sources of information: [5,11,12,14,17,28,29,30,31,32,48,51,55,60,61,62,63,64,65,66,67,68,69]

**Table 3.** Checklist of the Analysis Dimensions for RAG techniques

"N" = Naïve RAG; "C" = Corrective RAG; "G" = Graph RAG, "A" = Agentic RAG; "M" = Multimodal RAG; "S" = Self-RAG; "V" = Vision RAG; "F" = Fusion RAG; "R" = Radio RAG; "H" = HyPA RAG.
**Rating**: "vH " = Very High; "H" = High; "M" = Medium; "L" = Low; "vL" = Very Low; "L-M" = Low to Medium "M-H" = Medium to High; "-" = Insufficient information

| Dimension/ Metrics | Sub-dimensions/ | N | C | G | A | M | S | V | F | R | H |
|---|---|---|---|---|---|---|---|---|---|---|---|
| *Performance* | *Efficiency* | vH | M | L-M | vL | L-M | L | L-M | L | L-M | M |
| | *Scalability* | L | H | vH | vH | H | H | M | M | H | H |
| | *Retrieval Quality* | L | H | vH | vH | H | vH | H | M | M-H | vH |
| *Ethics* | *Transparency* | vH | - | vH | - | - | - | - | - | - | - |
| | *Fairness & Bias* | - | - | - | - | - | - | - | - | - | - |
| | *Accountability* | - | - | - | - | - | vH | - | - | - | - |
| *Trustworthiness* | *Correctness* | M | H | H | vH | M | H | H | M | H | H |
| | *Comprehensiveness* | L | M | vH | vH | H | M | M | H | M | H |
| | *Faithfulness* | M | vH | H | H | M | vH | H | M | H | H |
| | *Relevance* | M | H | H | vH | M | H | H | H | H | H |

| | | | | | | | | | | | | |
|---|---|---|---|---|---|---|---|---|---|---|---|---|
| *Safety* | | *Factuality* | L | H | vH | H | M | H | M | H | M | M-H |
| | | *Robustness* | L | M | H | M-H | L-M | H | L-M | H | M | M-H |
| | | *Privacy* | L | M | M | L-M | L | M | L-M | L-M | M | M-H |
| *Implementation Cost* | *Money Cost* | *Infrastructure* | vL | M-H | vH | H | vH | M | vH | M | M | H |
| | | *Developmental* | vL | M-H | H | vH | vH | H | H | M | M | H |
| | | *Operational* | vL | M-H | M | vH | H | H | H | M-H | M | H |
| | *Economic Efficiency* | *Cost Effectiveness Ration* | vH | M | L | vL | vL | M-H | L | M | M | H |
| | | *Retrieval Precision ROI* | L | H | vH | vH | M-H | H | M | H | M-H | H |
| | | *Cost Accuracy Trade-off* | vL | M | H | vH | L-M | H | L | M | H | H |
| | | *Comparative Cost Analysis* | - | - | - | - | - | - | - | - | - | - |
| Sources of information: [5,13,14,17,28,29,30,32,48,51,60,64,65,67,68] | | | | | | | | | | | | |

### 5.2 A Scenario Example

Consider the example of a set of tasks and user requirements for the RAG-LLM architecture of an Enterprise Financial and Legal Advisor (RAG-EFLA), as shown in Table 4. The checklists for the characterisation and analysis dimensions (Tables 2 and 3) can be used to comparatively assess which RAG techniques are the high-fit candidates that can meet the task and user requirements of the RAG-EFLA. The emphasis on multimodal integration (R3) and Agentic Intelligence (R5) makes Naïve RAG unsuitable. The likely candidates, based on Table 2 (characterisation checklist), are Agentic RAG, multimodal RAG, or Vision RAG. Also, regarding the RAG-EFLA requirements related to the analysis dimensions (Table 3), the emphasis on trustworthiness (R7), safety (R8 and R9), and efficiency (R11) will favour Graph RAG or Corrective RAG, which have high ratings in these categories. However, the implementation cost (R12), comprising monetary cost and economic efficiency, will favour RAG techniques with high ratings in these aspects. All of these considerations will determine the optimal RAG techniques for addressing the requirements of RAG-EFLA. In all cases, a weighted aggregate score can be computed for each RAG technique.

Given that a characterisation dimension rating could be:√ (Yes) = 1; † (partial) = 0.5; X (No) = 0; and
An analysis dimension rating could be: vH (Very High) = 5, H (High) = 4, M (Medium) = 3, L (Low) = 2, vL (Very Low) = 1, and Insufficient Information (-) = 0; L-M (Low to Medium) = 2.5 and M-H (Medium to High) = 3.5

The aggregate weighted score G can be computed as:

$$G = \sum_{i=1}^{M} W1_i C_i + \sum_{K=1}^{N} W2_k A_k \tag{1}$$

Where**:**

- $G$ is the total aggregate score for the evaluated RAG technique.
- $M$ is the total number of characterisation dimensions (e.g., M = 6).
- $N$ is the total number of analysis dimensions (e.g., N = 5).
- $W1_i$ is the user-assigned weight for the $i$-th characterisation dimension, where $1 \leq i \leq M$.
- $W2_k$ is the user-assigned weight for the $k$-th analysis dimension, where $1 \leq k \leq N$.
- $C_i$ is the aggregated rating score of the $i$-th characterisation dimension (calculated as the sum of its subdimensions and metrics).

- $A_k$ is the aggregated rating score of the $k$-th analysis dimension (calculated as the sum of its subdimensions and metrics).
- $C_i$ and $A_k$ are normalised average scores ranging from 0 to 1, since the characterisation and analysis dimensions have varying numbers of sub-dimensions or metrics.

$$C_i = \frac{1}{J_i} \sum_{j=1}^{J_i} f(r_{ij}) \tag{2}$$

Where:

- $J_i$ is the total number of metrics/sub-dimensions evaluated under characterisation dimension $i$.
- $r_{ij}$ is the qualitative symbol rating assigned to the $j$-th metric of the $i$-th dimension.

$$A_k = \frac{1}{5.P_k} \sum_{p=1}^{P_k} g(r_{kp}) \tag{3}$$

Where**:**

- $P_k$ is the total number of metrics evaluated under analysis dimension $k$.
- $r_{kp}$ is the linguistic rating assigned to the $p$-th metric of the $k$-th dimension.
- The constant 5 in the denominator scales the maximum possible Likert score (vH = 5) down to 1.

If the user-assigned weights are constrained to sum to 1 ($\sum W1_i + \sum W2_k$), the aggregate weighted score ($G$) will always be between 0 and 1, or 0% and 100%, providing a basis for comparing different RAG techniques to assess their potential fitness to meet a set of tasks and user requirements.

**Table 4**. Requirements of RAG-LLM Enterprise Financial and Legal Advisor (RAG-EFLA)

| id | Requirements | Category |
|---|---|---|
| These requirements define what the system must do and how it should behave structurally. They relate directly to the characterisation dimensions. | | |
| R1 | The system must perform both complex question answering and broad text summarisation (e.g., summarising a 200-page loan agreement and highlighting compliance risks into a user-prescribed maximum number of words). | Application Task & Language Generation |
| R2 | The assistant must navigate multiple domains simultaneously, seamlessly blending specialised financial accounting data with legal regulatory frameworks. | Domain Scope |
| R3 | Users must be able to upload PDF documents containing text, embedded financial tables, and performance charts. The system must extract and reason across both text and visual elements. | Multimodal Integration |
| R4 | To handle highly specific legal jargon alongside numeric financial metrics, the system must utilise a hybrid retrieval mechanism (combining sparse keyword matching for exact legal clauses with dense vector retrieval for semantic meaning). | Retrieval & Embedding Strategy |
| R5 | The assistant must execute multi-step reasoning, such as cross-referencing an uploaded contract against a live database of regulatory updates, identifying discrepancies, and formulating a corrected summary. | Agentic Intelligence |
| R6 | The system must adapt its output formatting and technical complexity based on user roles (e.g., providing high-level summaries for executive stakeholders vs detailed, granular citations for legal compliance officers). | Personalization & Adaptation |
| | These requirements define the performance boundaries, guardrails, and economic constraints of the system. They relate directly to the analysis dimensions. | |
| R7 | The system must maintain near-zero tolerance for hallucinations. Every generated assertion must be strictly grounded in the retrieved context with explicit citations. | Trustworthiness (Faithfulness and Factuality) |
| R8 | The system must gracefully handle ambiguous user prompts or conflicting information within documents without generating misleading advice or bypassing established safety guardrails. | Robustness and Safety |
| R9 | The system must strictly adhere to data privacy boundaries, ensuring that financial data from one department is not leaked or made visible to unauthorised users during retrieval. | Safety: Privacy and Compliance |
| R10 | The system must achieve very high retrieval precision to avoid cluttering the context window with irrelevant legal clauses.<br><br>The architecture must scale efficiently across millions of enterprise documents without suffering a drop in retrieval accuracy or speed. | Retrieval Quality and Performance |
| R11 | Response times for standard question answering must remain under 3 seconds (High Efficiency), even when executing agentic reflection loops. | Efficiency |
| R12 | The development and operational infrastructure costs must balance accuracy with financial viability. While High correctness is required, it must avoid runaway token costs from brute-force agent loops. | Economic Cost and Implementation |

Based on the RAG-EFLA requirements (R1-R12), ratings can be assigned to different RAG techniques according to the characterisation and analysis dimensions checklists (See Tables 2 and 3). The ratings after applying the characterisation dimensions and analysis dimensions are shown in Tables 5 and 6.

**Table 5**: Application of Characterisation Dimension

"N" = Naïve RAG; "C" = Corrective RAG; "G" = Graph RAG; "A" = Agentic RAG;
"M" = Multimodal RAG; "S" = Self-RAG; "V" = Vision RAG; "F" = Fusion RAG; "R" = Radio RAG; "H" = HyPA RAG.
√ (Yes) = 1; † (partial) = 0.5; X (No) = 0; Number of dimensions/subdimensions = 12

| ReqID | Dimension /Subdimensions/ Metrics | | | N | C | G | A | M | S | V | F | R | H |
|---|---|---|---|---|---|---|---|---|---|---|---|---|---|
| R4 | *Retrieval Mechanism* | | *Sparse Retrieval* | √ | √ | √ | √ | √ | √ | √ | √ | √ | √ |
| | | | *Dense Retrieval* | √ | √ | √ | √ | √ | √ | √ | √ | √ | √ |
| | | | *Hybrid Retrieval* | √ | √ | √ | √ | √ | √ | √ | √ | √ | √ |
| R5 | *Agentic Intelligence* | | | X | √ | † | √ | † | √ | † | X | X | √ |
| R2 | *Applic ation* | *Domain* | *Single* | √ | √ | √ | √ | √ | √ | √ | √ | √ | √ |
| | | | *Multiple* | X | √ | √ | √ | √ | √ | √ | √ | X | √ |
| R1 | | *Task* | *Text Summarisation* | † | √ | √ | √ | √ | √ | † | √ | √ | √ |
| | | | *Language Generation* | √ | √ | √ | √ | √ | √ | √ | √ | √ | √ |
| R4 | *Multimodal Integration* | | *Embedding Techniques* | † | √ | √ | √ | √ | X | √ | √ | X | √ |
| R3 | | | *Context-to-Textual* | X | √ | √ | √ | √ | √ | † | √ | √ | √ |

| | | | N | C | G | A | M | S | V | F | R | H |
|---|---|---|---|---|---|---|---|---|---|---|---|---|
| R6 | *Personalization & Adaptation* | *Personalisation* | X | √ | √ | √ | † | X | † | X | X | X |
| | | *Adaptation* | X | √ | √ | √ | √ | √ | √ | √ | √ | √ |
| Sum of ratings (CD) | | | 8 | 12 | 11.5 | 12 | 11 | 10 | 10 | 10 | 8 | 11 |
| Normalised scores ($C_i$ = CD/12) | | | 0.66 | 1 | 0.95 | 1 | 0.91 | 0.83 | 0.83 | 0.83 | 0.66 | 0.91 |

**Table 6**: Application of Analysis Dimension

"N" = Naïve RAG; "C" = Corrective RAG; "G" = Graph RAG, "A" = Agentic RAG; "M" = Multimodal RAG; "S" = Self-RAG; "V" = Vision RAG; "F" = Fusion RAG; "R" = Radio RAG; "H" = HyPA RAG. "-" = Insufficient information
**Ratings**: vH=5; H=4; M-H=3.5; M=3; L-M=2.5; L= 2; vL=1. Number of dimensions/subdimensions = 9

| REQID | Dimension/ Sub-dimensions/ Metrics | | | N | C | G | A | M | S | V | F | R | H |
|---|---|---|---|---|---|---|---|---|---|---|---|---|---|
| R11 | *Performance* | | *Efficiency* | vH | M | L-M | vL | L-M | L | L-M | L | L-M | M |
| R10 | | | *Retrieval Quality* | L | H | vH | vH | H | vH | H | M | M-H | vH |
| R7 | *Trustworthiness* | | *Faithfulness* | M | vH | H | H | M | vH | H | M | H | H |
| R7 | *Safety* | | *Factuality* | L | H | vH | H | M | H | M | H | M | M-H |
| R8 | | | *Robustness* | L | M | H | M-H | L-M | H | L-M | H | M | M-H |
| R9 | | | *Privacy* | L | M | M | L-M | L | M | L-M | L-M | M | M-H |
| R12 | *Implementation* | *Economic Efficiency* | *Cost Effectiveness Ratio* | vH | M | L | vL | vL | M-H | L | M | M | H |
| R12 | | | *Development al Cost* | vL | M-H | H | vH | vH | H | H | M | M | H |
| R12 | | | *Operational Cost* | vL | M-H | M | vH | H | H | H | M-H | M | H |
| Sum of ratings (AD) | | | | 25 | 32 | 33.5 | 31 | 27 | 34.5 | 28.5 | 28 | 28 | 34.5 |
| Normalised scores ($A_k$ = AD/(5*9)) | | | | 0.55 | 0.71 | 0.74 | 0.68 | 0.6 | 0.77 | 0.63 | 0.62 | 0.62 | 0.76 |

Based on the normalised rating scores, the Aggregate Score (G) for specific RAG techniques (e.g. NaiveRAG (N), AgenticRAG (R), etc.) can be computed using the formula in equation (1) as follows:

**Case 1: all the dimensions (characterisation and analysis) have equal weights that sum up to 1**

$$\sum_{i=1}^{12} W1_i + \sum_{K=1}^{9} W2_k = 1 \qquad (4)$$

21*W = 1, the W=1/21 = 0.0476.
Total weight for characterisation dimensions (W1) = 12/21 = 4/7 = 0.57
Total weight for analysis dimension (W2) = 9/21 = 3/7 = 0.43
The formula for computing G then becomes:

$$G = (0.57 \times C_i) + (0.43 \times A_k)$$

Naïve RAG: $G(N) = (0.57 \times 0.66) + (0.43 \times 0.55) = 0.6127 \Rightarrow 61.27\%$
Corrective RAG: $G(C) = (0.57 \times 1) + (0.43 \times 0.71) = 0.8753 \Rightarrow \mathbf{87.53}\%$
Graph RAG: $G(G) = (0.57 \times 0.95) + (0.43 \times 0.71) = 0.8468 \Rightarrow \mathbf{84.68}\%$
Agentic RAG: $G(A) = (0.57 \times 1) + (0.43 \times 0.74) = 0.8882 \Rightarrow \mathbf{88.82}\%$
Multimodal RAG: $G(M) = (0.57 \times 0.91) + (0.43 \times 0.68) = 0.8111 \Rightarrow 81.11\%$
Self RAG: $G(S) = (0.57 \times 0.83) + (0.43 \times 0.6) = 0.7311 \Rightarrow 73.11\%$
Vision RAG: $G(V) = (0.57 \times 0.83) + (0.43 \times 0.77) = 0.8042 \Rightarrow 80.42\%$
Fusion RAG: $G(F) = (0.57 \times 0.83) + (0.43 \times 0.66) = 0.7569 \Rightarrow 75.69\%$
Radio RAG: $G(R) = (0.57 \times 0.66) + (0.43 \times 0.62) = 0.6428 \Rightarrow 64.28\%$

HyPA RAG: $G(H) = (0.57 \times 0.91) + (0.43 \times 0.76) = 0.8455 \Rightarrow \mathbf{84.55}\%$

Based on the aggregate scores, the best high-fit RAG candidates for the RAG-EFLA requirements are Agentic RAG, Corrective RAG, Graph RAG, and HyPA RAG. The best RAG technique can subsequently be determined empirically using RAG benchmarking. However, the pool of 10 RAG techniques is pruned to 4 high-fit candidates using checklists derived from the proposed conceptual framework.

**Case 2: Experts assign different custom weights to dimensions, but all weights sum up to 1**.

Practically, experts will likely assign a macro-weight to each category, for example, 60% to characterisation dimensions (W1), and 40% to analysis dimensions (W2). The set of sub-weights ($w1_i$ and $w2_k$) will then be distributed within each group.

$$G = W1 \sum_{i=1}^{M} w_{1k} C_i + W2 \sum_{1}^{N} w_{2k} A_k \qquad (5)$$

Where

$W_1 + W_2 = 1$ (e.g. $W_1 = 0.6$ for Characterisation, $W_2 = 0.4$ for Analysis)

$\sum_{i=1}^{M} w_{1i} = 1$ (the internal sub-weights of Characterisation dimensions that sum to 1)

$\sum_{k=1}^{N} w_{2i} = 1$ (the internal sub-weights of Analysis dimensions that sum to 1)

Since M = 12 and N = 9

If the experts assign macro weights 0.6 (characterisation) and 0.4 (Analysis), and also assign subweights to individual dimensions within each group, the formula in (5) to compute the total aggregate score then becomes:

$$= 0.6 \sum_{i=1}^{12} w_{1k} C_i + 0.4 \sum_{1}^{9} w_{2k} A_k \qquad (6)$$

By applying the formula in (6), the aggregate score (G) for different RAG techniques can be computed to determine the high-fit RAG candidates.

## 6. Conclusion

In this study, we proposed a conceptual framework comprising dimensions, subdimensions, and concepts/metrics for comparatively assessing how well different RAG techniques satisfy specific task and user requirements. The conceptual framework can be used as a checklist and filtering mechanism to identify high-fit RAG candidates that meet a set of tasks and user requirements before empirical validation. Thus, the conceptual framework is a pre-empirical assessment tool for RAG techniques that could be used to alleviate the huge cost associated with empirical benchmarking of RAG techniques, which can be a computationally and financially expensive procedure.

To achieve this, we employed an iterative seven-step process that involves selecting relevant papers from the literature: paper selection; determining the key themes; iv) identifying dimensions; v) identifying sub-dimensions; vi) identifying characteristics and metrics of dimensions/sub-dimensions; and quality assurance. So far, no classification of RAG techniques by architectural and operational attributes has been proposed in the literature, which limits understanding of their intrinsic composition, features, and capabilities. Thus, as a contribution, this paper proposes a comprehensive conceptual framework that encompasses multiple critical dimensions to enable rigorous characterisation and analysis of the features and capabilities of RAG techniques, providing a robust basis for decision-making when selecting RAG techniques that fit specific tasks and user requirements before conducting empirical benchmarking.

A major limitation of this study is that the framework is only applied to a set of selected (the more popular) RAG techniques, and not to all the available RAG techniques reported in the literature. The plausibility of our conceptual framework is that it can be applied to new RAG techniques once additional data has been collected from the literature. Another limitation is that the assigned ratings were not based on empirical validation; rather, they were a product of inferences from the information reported in the literature. Since there is no guarantee of the absolute correctness of the facts reported on specific RAG techniques, our study inherits the reliability, errors, and inaccuracies that may pertain to the papers used as sources of information for our analysis.

In future work, we will apply the conceptual framework to conduct a comparative evaluation of the features and capabilities of different RAG techniques across use cases, further validating its plausibility. We will also conduct empirical RAG benchmarking to validate the rating assigned to specific dimensions and sub-dimensions of the conceptual framework. Specific attention will also be given to the ethical attributes of RAG implementations, for which the literature currently provides insufficient information.